# TumorXAI: Self-Supervised Deep Learning Framework for Explainable Brain MRI Tumor Classification

Abrar Hossain Zahin[1], Amit Kumar Saha[1], Tanvir Mridha[1], Saifur Rahman[1], Jannatul Ferdous Prome[1], Raima Husna[1], Israt Jahan[1], Ahmed Wasif Reza[1*]

[1]Department of Computer Science and Engineering, East West University, Dhaka, 1212, Bangladesh.

Contributing authors: abrarhossain1200@gmail.com; amitkumarsahaak3@gmail.com; tanvirmridha17540042@gmail.com; saifur3734@gmail.com; prome128900@gmail.com; raimahusna100@gmail.com; isratjahan2744829@gmail.com; wasif@ewubd.edu;

**Abstract**

Classifying brain tumors using magnetic resonance imaging (MRI) is crucial for early diagnosis and treatment; however, tumor heterogeneity and a dearth of annotated datasets restrict the use of supervised deep learning approaches. In this work, we use self-supervised learning (SSL) to study multi-class brain tumor classification. Using a ResNet-50 backbone, we evaluate four SSL frameworks including SimCLR, BYOL, DINO, and Moco v3 on a publicly available dataset of 4,448 MRIs with 17 distinct tumor types. On the dataset, SimCLR achieved **99.64%** accuracy, 99.64% precision, 99.64% recall, and 99.64% F1-score. The workflow includes preprocessing, fine-tuning, linear evaluation, and SSL pretraining with data augmentations. Results show that, when labels are limited, SSL-pretrained models outperform supervised baselines in terms of F1-score, recall, accuracy, and precision. Additionally, by providing visual insights into model decisions, Explainable AI techniques (Grad-CAM, Grad-CAM++, EigenCAM) enhance interpretability. These results demonstrate SSL's scalability and dependability in diagnosing brain tumors from unlabeled medical data.

## 1 Introduction

Brain tumors are highly complex and heterogeneous and rank among the most lethal neurological diseases, accounting for considerable cancer morbidity and mortality. These are becoming more common, and detecting the same early is crucial for survival uplift. Although Magnetic Resonance Imaging (MRI) is the most effective modality that provides more anatomical information and is non-ionizing, radiologists still face challenges interpreting MRI images due to the heterogeneity of tumors, with adjacent tumors overlapping [1]. This complexity also leads to an investigation that focused on artificial intelligence (AI), deep learning (DL) to facilitate the automatic brain tumor classification in order to improve the diagnosis precision.

Over the years, Convolutional Neural Networks (CNNs) have yielded remarkable results in medical imaging [2]. But these models rely on large-scale labelled datasets that are expensive and time-consuming to obtain, and do not apply to many medical problems, because some tumor types are rare, hence class imbalance. To overcome these drawbacks, self-supervised learning (SSL) has proven to be a potent alternative, where pretext tasks leverage the abundant unlabeled data to learn useful feature representations. Deep learning based self-supervised learning (SSL) techniques such as SimCLR, BYOL, and DINO have reached a new SOTA for image-level representation learning without explicitly requiring labels as supervision [3], but have been relatively less explored for complex medical tasks (e.g. brain tumor classification).

The goal of this investigation is to measure memory-based multi-class brain tumor classification performance levels over an MRI image dataset with different abnormalities. Based on the ResNet-50 architecture, the research uses three popular SSL frameworks for pretraining: SimCLR, BYOL, and DINO. The pipeline consists of systematic dataset preprocessing followed by data augmentation, SSL-based feature learning, linear evaluation, and fine-tuning in the context of 17 tumor types. It also includes Explainability AI (XAI) methods like Grad-CAM for providing model explainability and interpretability [4], which are critically important for clinical practices.

Evaluation of 4,448 MRI images shows that these feature representations of images are very strong, semantic, and relevant if the model is pre-trained with SSL methods [5]. Experimental results show that explicit SSL methods outperform classic supervised approaches, even when a small amount of labeled data is available. Visualization techniques confirm that SSL models produce a well-formed feature space that enhances tumor class separation. Moreover, the interpretability methods strengthen AI prediction in clinical relevance.

To the first and best of our knowledge, research complementing the global debate on self-supervised learning in medical image analysis, our study provides an exhaustive comparative analysis of SSL methods for multi-class brain tumor classification. These results showcase the ability of SSL to alleviate data scarcity, improve diagnostic performance, and increase model interpretability, which, together, make SSL a new scalable and trusted tool for assisting radiologists in making efficient and accurate clinical decisions [6].

The remainder of this paper is organized as follows: Section 2 reviews related works. Section 3 illustrates the datasets used, the preprocessing techniques applied, and the deep learning models implemented in this study. Section 4 represents the experimental results, comparative performance analysis, and Grad-CAM visualizations. Section 5 describes the findings, limitations, and practical implications of the proposed approach. Finally, the paper concludes and highlights potential future research directions.

## 2 Related Works

The rising incidence of cotton leaf diseases has prompted an increasing reliance on deep learning methodology for timely and accurate diagnosis. A variety of convolutional neural networks (CNNs) and transfer learning strategies have been applied to this domain, each offering varying degrees of accuracy, generalizability, and deployment feasibility [7].

Several studies have applied traditional deep learning frameworks. Nagarjun et al. [8], incorporating ResNet101, InceptionV2, DenseNet121, and a custom-designed CNN, were evaluated on a dataset comprising 2,616 images. Among these, the custom CNN achieved an impressive (99%) accuracy. However, the model's effectiveness was potentially limited by the modest dataset size and evident class imbalance, which could undermine its performance in uncontrolled environments. To broaden this scope, a separate study employed a larger dataset of 6,158 images across eight categories, testing multiple pre-trained models. VGG16 emerged as the best performer, attaining a classification accuracy of (95.02%). Yet, the presence of visually similar disease classes and diverse image conditions posed challenges to consistent robustness.

Another investigation, Arshleen Kaur et al. [9], involved fine-tuning the VGG16 model using a dataset augmented to 5,000 images. This approach yielded (95.5%) accuracy, demonstrating sensitivity to training configurations and suggesting limitations in broader generalizability. Similarly, Azath M. et al. [10] analyzed InceptionV3, MobileNetV1/V2, and Xception was conducted on a relatively small dataset of 2,258 images, with Xception achieving a peak accuracy of 98.61%. Nonetheless, the authors advocated for further experimentation in field conditions to account for environmental variations.

Himani Bhatheja et al.[11] explored using multiple architectures, where a custom CNN outperformed others by exceeding (98%) accuracy. However, this was achieved using only 106 images, prompting concerns over scalability and deployment feasibility in large-scale agricultural settings. R. Sharma et al. [12] A VGG16-based model trained on 10,000 images targeting four cotton leaf diseases achieved a respectable (93.8%) accuracy. Despite this, reliance on publicly sourced datasets and moderate test set sizes may have affected its adaptability to field conditions.

S. Bavaskar et al. [13]. applied various CNN models on a balanced dataset of 1,711 images representing four disease categories. Here, ResNet152V2 recorded an accuracy above 98%, though dependence on curated Kaggle datasets raised questions about generalization beyond controlled settings. In another instance, Mrs. C. M. Devi et al. [14] used CNNs with optional transfer learning on mixed-source image datasets, achieving promising results. Yet, performance remained tied to web-sourced images, necessitating real-world validation.

A broader assessment of deep learning and traditional algorithms using large public datasets like PlantVillage demonstrated CNN dominance, with accuracies ranging between (98–99%). Still, the absence of deployment-focused measures and heavy reliance on curated data remained critical gaps H. M. Faisal et al. [18]. Additional work incorporated Fourier and wavelet transforms into CNNs such as VGG19 and InceptionV3, attaining (98.4%) classification accuracy. However, issues like insufficient disease diversity and lack of deployable platforms persisted, V. Burkpalli et al [15].

In another hybrid study, A. Kumar et al.[16] image-based models (VGG16, ResNet50) were fused with IoT-generated sensor data, yielding a high AUC of 0.9965. Nevertheless, the dataset's single-source origin reduced its adaptability to other agricultural regions. A machine learning ensemble integrating Random Forest, SVM, and additional techniques attained a near (98%) accuracy. Still, its focus on region-specific data and the absence of explainable AI approaches limited its real-world relevance, G. Singh et al.[17].

Efforts to combine DenseNet121 and ResNet50 in a hybrid CNN model achieved (96%) accuracy across six disease categories. However, synthetic data biases and a lack of field trials raised concerns about overfitting Gunjan Gupta et al[18]. Similarly, an ensemble of VGG16 and InceptionV3 showed strong results(98%) training and (95%) testing accuracy, but faced limitations due to dataset narrowness and no in-field validation, Azath M. et al.[9]. We summarize A common limitation in cotton disease detection studies based on deep learning is the class imbalance, which leads to biased models that do not generalize well to real-world scenarios.

Here, we report successful advances in the detection of brain tumors by using artificial intelligence (AI) and deep learning (DL) techniques. Rahman et al. [17] An intelligent real-time tumor detection system using IoT was developed by (2022). Faisal et al. [18] Background: Computer-assisted diagnostic systems improve the precision of reported imaging results (2013). Reza et al. [20] developed a CNN-based approach for brain tumor classification from MRI (2023). Additionally, Siddiqui et al. [22] Sara Jennifera et al.(2005) [19]discussed explainable artificial intelligence-based stacking ensemble methods for cervical cancer diagnosis. Classification of Sickle Cell Disease Using State-of-the-Art Deep Learning Models. (2023) All these studies highlight the important role of AI and deep learning in medical diagnosis.

The latest developments of deep learning have shown to enhance diagnostic systems for humans and diseases in agriculture. Rahman et al. (2022) proposed an IoT-oriented brain tumour detection system. Faisal et al. (2013) developed a computerized algorithm for a tumor radiograph system. Reza et al. (2023) Keras relu and Keras ReLU developed a CNN model for MRI-based brain tumor classification with MRI scans, and Sara Jennifer et al. (2023) used deep learning for the classification of individuals with sickle cell disease. In agriculture, Al et al. (2025) [23] detected mango leaf diseases using AI with explainable AI. These works demonstrate increasing contributions of deep learning and explainable AI (XAI) to enhancing disease diagnosis in multiple disciplines.

# 3 Methodology

Figure 1 depicts a sequential pipeline that begins with the acquisition and preparation of Brain Tumor MRI Images (17 classes),followed by normalizing, and splitting the dataset into 80% training, 10% each for validation and test. To enhance model robustness, we applied different SSL algorithms. Datasets → preprocessing → model training → linear eval → fine-tune → testing with SSL models to evaluate metrics which consists of accuracy, precision, recall and F1-score. Lastly, in order to improve model understandability, Explainable AI methods such as Grad- CAM, Grad-CAM++, Eigen-CAM are used.

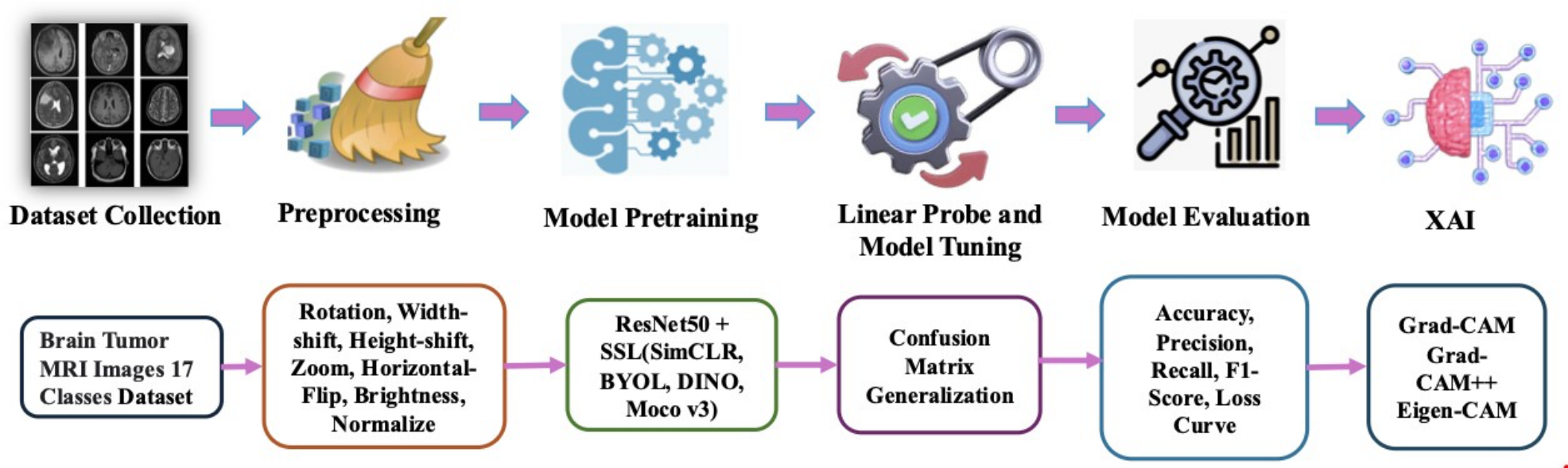

**Fig. 1** Workflow Diagram

## 3.1 Dataset Description

For this research, we utilize Brain Tumor MRI Images (17 Classes), openly available Kaggle data. The data contains MRI scans that are clustered into 17 types of tumor classes, offering a wide range of multi-class classification of tumors. The distribution of the dataset over varying classes of tumors is summarized in Table 1. The data exhibit a class real-world imbalance medical dataset, such that there are significantly fewer samples of certain types of tumors relative to others. Figure 2 shows a random image from the tumor class is visualized.

**Table 1** Summary of Brain Tumor MRI Dataset (17 Sub Classes).

| Tumor Class | Amount |
|---|---|
| Glioma (T1C+, T1, T2) | 508 + 430 + 346 = 1284 |
| Schwannoma (T1, T1C+, T2) | 153 + 194 + 123 = 470 |
| Meningioma (T1, T1C+, T2) | 345 + 625 + 329 = 1299 |
| Neurocitoma (T1, T1C+, T2) | 169 + 261 + 112 = 542 |
| Other Lesions (T1, T1C+, T2) | 152 + 48 + 57 = 257 |
| Normal (T1, T2) | 272 + 291 = 563 |
| **Total** | **4448** |

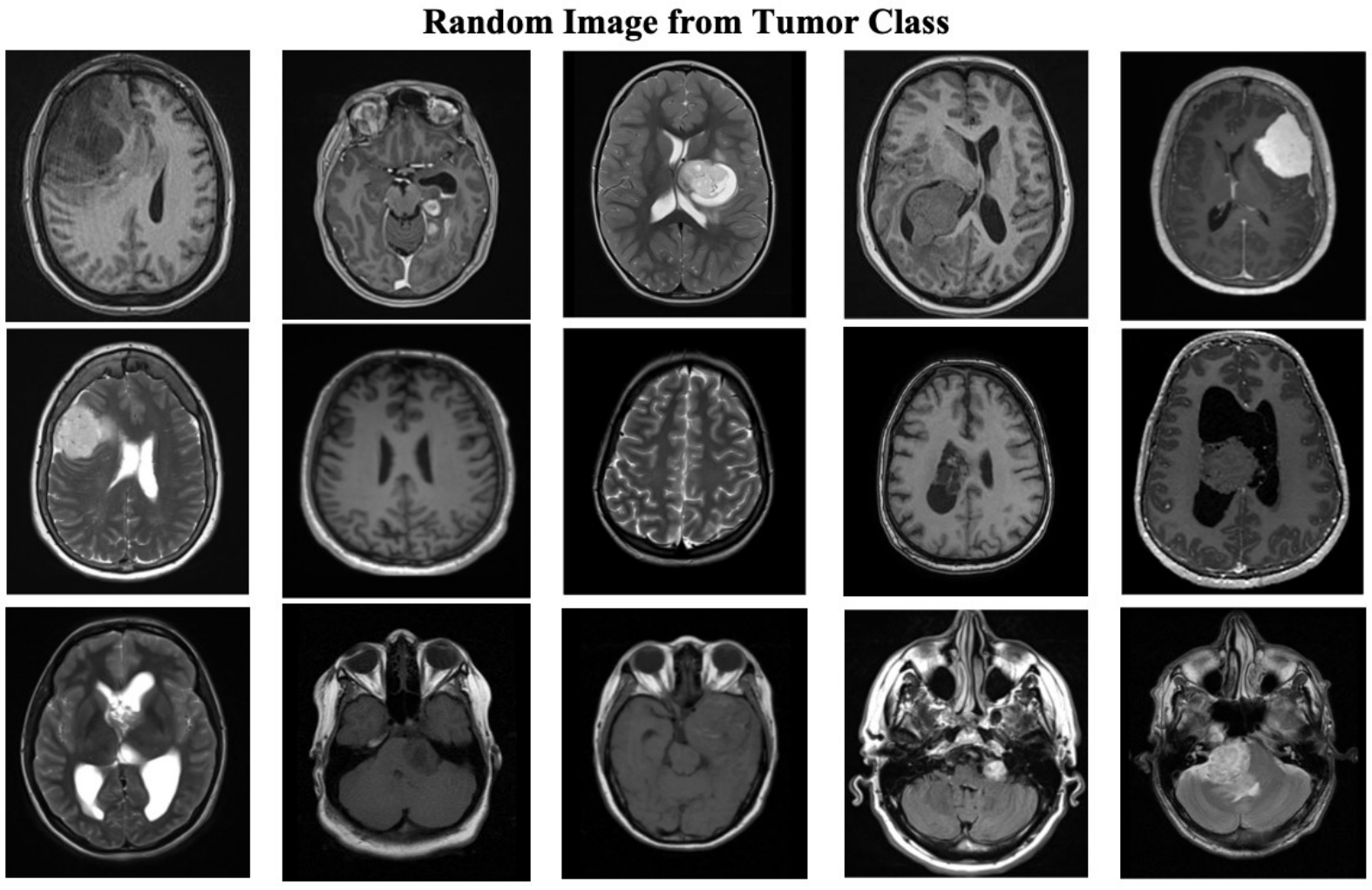

**Fig. 2** Visualization of a Random Image for Tumor Class.

## 3.2 Dataset Preprocessing

For self-supervised learning, strong data augmentations are required. In Table 2, random resizing and cropping, flipping, and rotation of the images horizontally, and adding a Gaussian blur to help the network learn robust features.

**Table 2** Image Data Augmentation Parameters Used.

| Parameter | Value |
|---|---|
| Rotation Range | Random rotation within ±20° |
| Width Shift Range | Random horizontal shift up to 10% of total width |
| Height Shift Range | Random vertical shift up to 10% of total height |
| Zoom Range | Random zoom within ±10% |
| Horizontal Flip | Random left-right flipping applied |
| Brightness Range | Adjust brightness within range [0.8,1.2] |
| Fill Mode | Nearest-neighbor interpolation |

Following the data augmentation process, the size of the dataset is upscaled and balanced at the station for each category (Handwritten/printed), augmented to 625 samples. This step decreased class imbalance, and we obtained a train (8500 images), valid (1062 images), and test set (1063 images). The enhancement increased the reliability of datasets and the generalization of models. Some randomly selected augmented examples are shown in Figure 3, and we can observe that the diversity and variations across classes are consistently increased, which can further lead to more robust models.

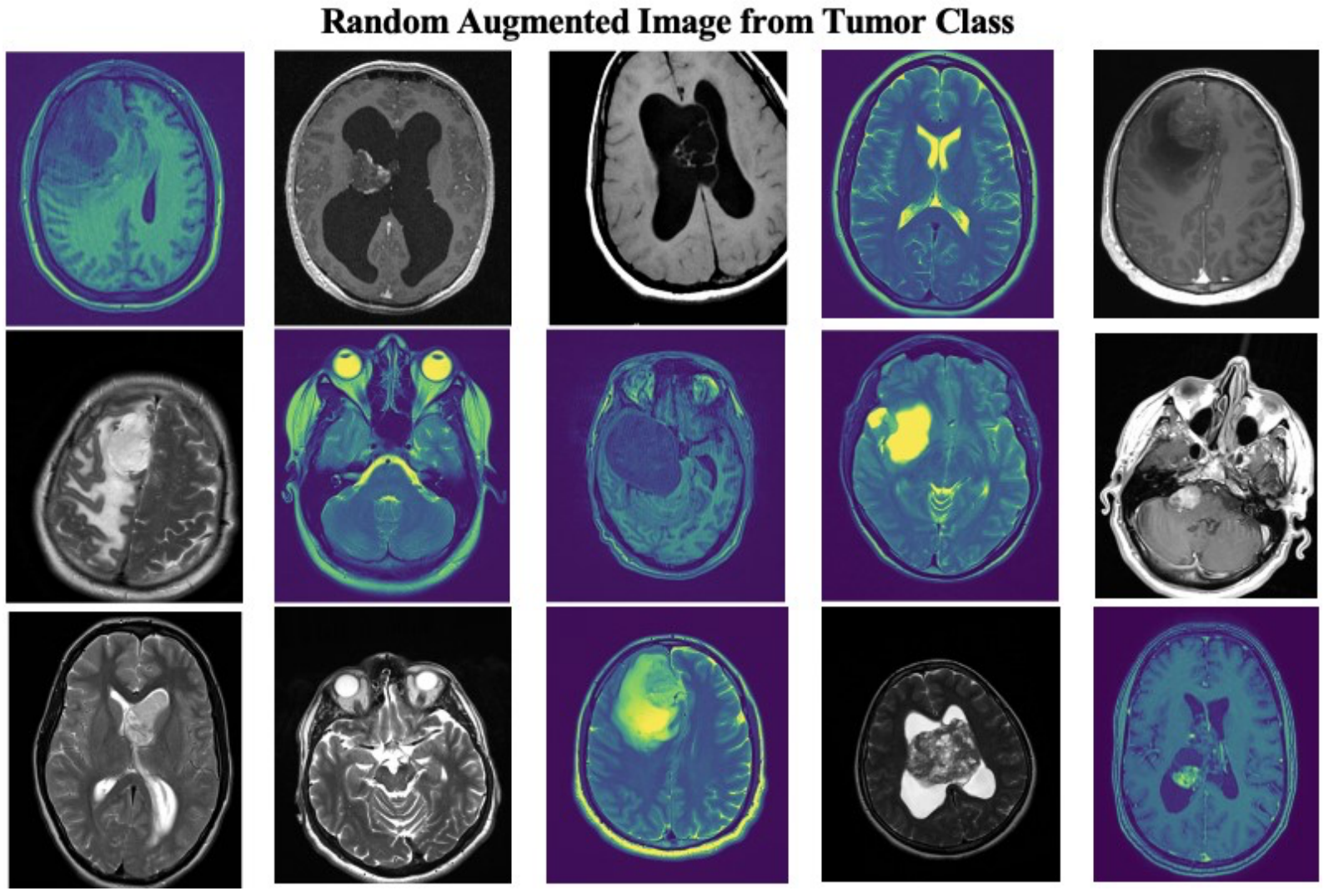

**Fig. 3** Visualize Random Augmented Image for Tumor class

## 3.3 Models and Implementation

### 3.3.1 SimCLR Self-Supervised Learning

We employ a standard ResNet-50 encoder for images with the final fully connected classification layer turned off. The encoder for an input image of size 224×224 produces a pooled global feature vector of 2048 size. During self-supervised pretraining, a lightweight projection head (contrastive head) is appended to the encoder.

$$\text{Linear}(2048 \rightarrow 512) \rightarrow \text{ReLU} \rightarrow \text{Linear}(512 \rightarrow 128) \tag{1}$$

$$\text{Data Augmentation:} \qquad \widetilde{x_i^{(1)}}, \widetilde{x_i^{(2)}} \sim T(\cdot \mid x_i) \tag{2}$$

Encoder and Projection Head:

$$h_i^{(a)} = f_{\theta}\backslash big\left(\widetilde{x_i^{(a)}}\backslash big\right), \quad z_i^{(a)} = g_{\phi}\backslash big\left(h_i^{(a)}\backslash big\right), \quad a \in \{1,2\} \tag{3}$$

$$\text{Normalization:} \qquad \bar{z} \equiv \frac{z}{\|z\|_2} \tag{4}$$

$$\text{Similarity Metric:} \qquad \text{sim}(u, v) = u^{\top} v \tag{5}$$

$$\text{NT} - \text{Xent Loss:} \qquad \text{l}_i^{(a)} = -\log \frac{\exp\left(\text{sim}\left(\overline{z_i^{(a)}}, \overline{z_i^{(b)}}\right)/\tau\right)}{\sum_{k=1,c=1,(k,c)\neq(i,a)}^{N,2} \exp\left(\text{sim}\left(\overline{z_i^{(a)}}, \overline{z_k^{(c)}}\right)/\tau\right)} \tag{6}$$

$$\text{Total SimCLR Loss:} \qquad \mathcal{L}_{\mathcal{S}m\mathcal{L}} = \frac{1}{2N} \sum_{i=1}^{N} \left(\text{l}_i^{(1)} + \text{l}_i^{(2)}\right) \tag{7}$$

Similarity Matrix: $$S = \frac{1}{\tau} ZZ^{\top} \quad (8)$$

Simplified Log-Loss $$l_i = -\log \frac{\exp(s_{i,j(i)})}{\sum_{k \neq i} \exp(s_{i,k})} \quad (9)$$

Figure 4 overview of the SimCLR model architecture for self-supervised representation learning, consisting the encoder backbone, projection head, and contrastive learning framework.

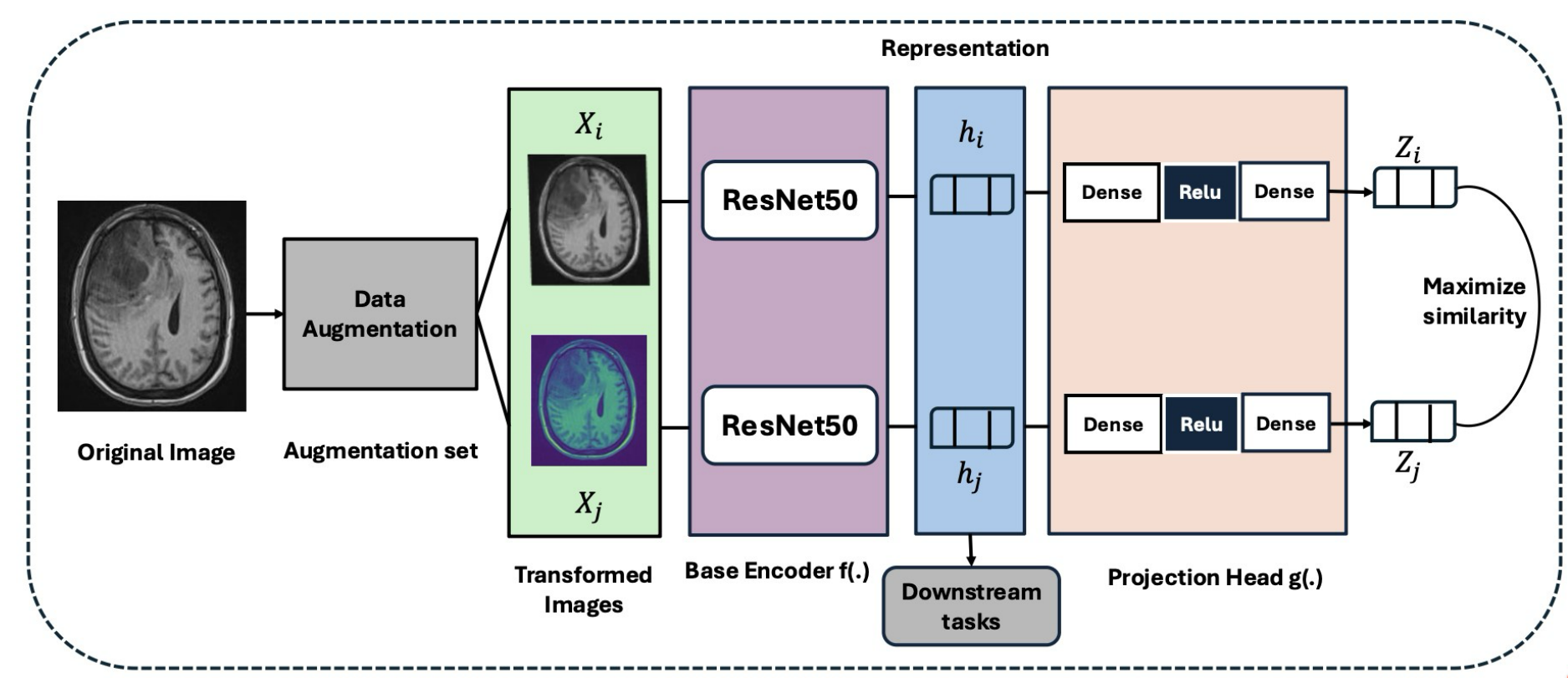

**Fig. 4** SimCLR self-supervised learning architecture

**ResNet50:** ResNet50: The ResNet is a deep CNN with 50 layers, which solves the degradation problem of deep neural networks by stacking the residual units. Each block is composed of three convolutions: 1×1 for dimension reduction, 3×3 for feature extraction, and 1×1 for dimension restoration. The network consists of a convolutional stem, four stages with increasingly large filters and stride-2 convolutions for downsampling. It is efficient and effective for image classification tasks and self-supervised learning.

The residual block is defined as:

$$\mathbf{y_l} = \mathcal{F}(\mathbf{x_l}, \{W_l\}) + \mathbf{x_l} \quad (10)$$

$$\mathbf{x_{l+1}} = f(\mathbf{y_l}) \quad (11)$$

The model is built by stacking 49 residual blocks, followed by a fully connected classification layer:

$$\hat{\mathbf{y}} = \mathrm{softmax}(W_{fc} \cdot \mathbf{z} + b_{fc}) \quad (12)$$

The architecture of the ResNet50 model has been shown in Figure 5 and is a simple backbone block for self-supervised learning.

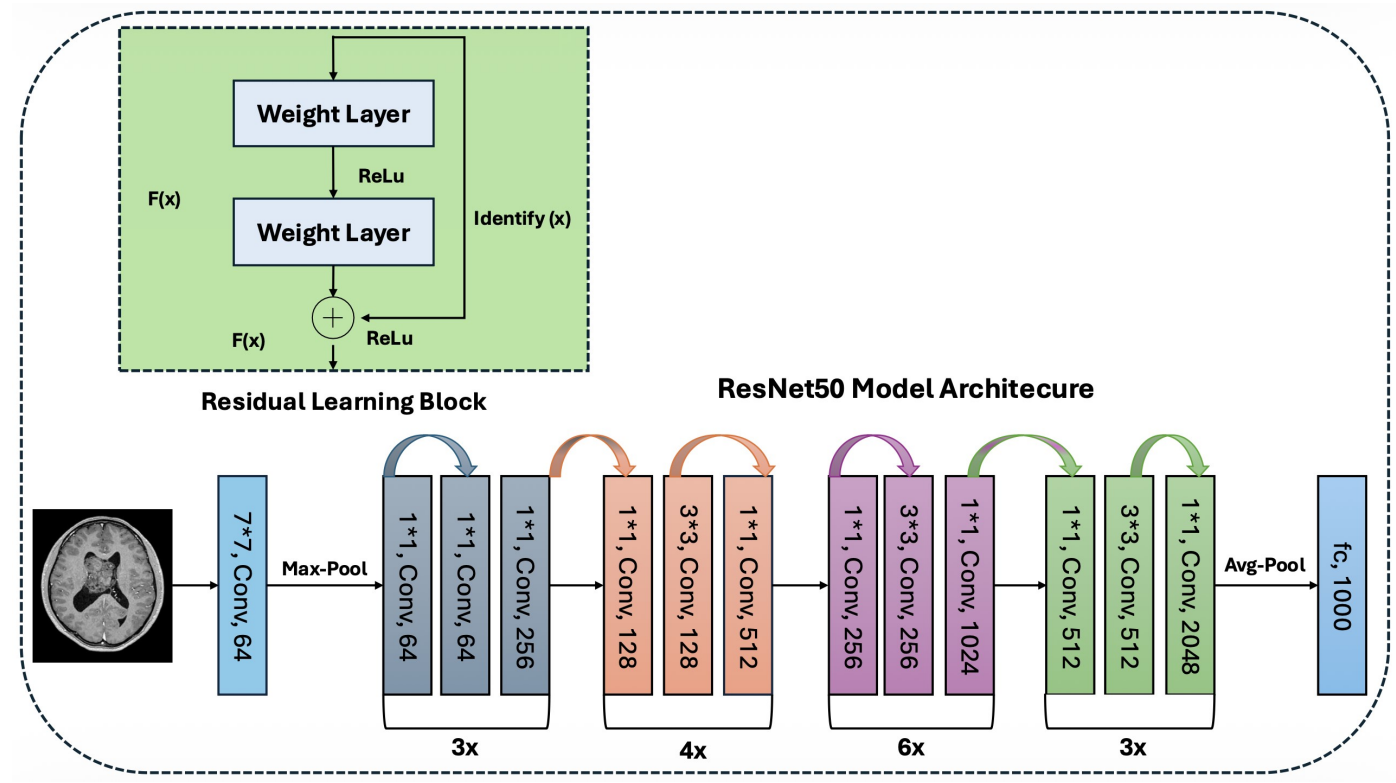

**Fig. 5** ResNet50 Architecture Diagram

## 3.4 Experimental Setup

Hyperparameters for training SSL methods with a ResNet50 encoder are given in Table 3. We employed AdamW as the optimizer with a weight decay equal to $1\times10^{-6}$ for DINO and SimCLR, and $1\times10^{-4}$ for MoCo v3 and BYOL. Learning rates were $1\times 10^{-3}$ for MoCo v3, $1\times10^{-4}$ for DINO and BYOL, and $3\times10^{-4}$ for SimCLR. Same for tuning and linear-probe rates: specific to methods. Temperature setting for MoCo v3 $\tau = 0.2$, DINO student/teacher temperature separate, SimCLR temperature $\tau = 0.5$. At the end of the model fitting, we used cross-entropy loss and early stopping (patience = 10). These consistent settings eliminate algorithmic differences. As shown in Table 3, the hyperparameters for MoCo v3, DINO, BYOL, and SimCLR are compared.

**Table 3** Comparative Hyperparameter Configuration of MoCo v3, DINO, BYOL, and SimCLR with ResNet50 Backbone on the Brain Tumor MRI Dataset

| **Hyperparameter** | **MoCov3** | **DINO** | **BYOL** | **SimCLR** |
|---|---|---|---|---|
| Batch Size | 32 | 32 | 32 | 32 |
| Number of Workers | 4 | 4 | 4 | 4 |
| Epochs (SSL Pretraining) | 80 | 80 | 80 | 80 |
| Epochs (Linear Evaluation) | 50 | 50 | 50 | 50 |
| Epochs (Fine-tuning) | 50 | 50 | 50 | 50 |
| Learning Rate (LR) (SSL) | $1 \times 10^{-3}$ | $1 \times 10^{-4}$ | $1 \times 10^{-4}$ | $3 \times 10^{-4}$ |
| LR (Linear Probe) | $3 \times 10^{-3}$ | $3 \times 10^{-3}$ | $1 \times 10^{-3}$ | $1 \times 10^{-3}$ |
| LR (Fine-tuning) | $5 \times 10^{-4}$ | $1 \times 10^{-4}$ | $1 \times 10^{-4}$ | $1 \times 10^{-4}\times 10$ |
| Optimizer | AdamW | AdamW | AdamW | AdamW |
| Temperature | 0.2 | Student = 0.1, Teacher = 0.04 | - | 0.5 |
| Weight Decay | $1 \times 10^{-4}$ | $1 \times 10^{-6}$ | $1 \times 10^{-4}$ | $1 \times 10^{-6}$ |
| Loss Function | CrossEntropy | CrossEntropy | CrossEntropy | CrossEntropy |
| EarlyStopping | patience=10 | patience=10 | patience=10 | patience=10 |

# 4 Result Analysis

## 4.1 Evaluation of the Model's Performance

As shown in Table 4, the four self-supervised learning models—SimCLR, BYOL, DINO and MoCo v3—perform comparably in fine-tuning and linear evaluation settings. In fine-tuning, SimCLR achieves an accuracy of 99.64% with perfect precision, recall, and F1 scores (0.9964). Then comes BYOL with an accuracy of 97.66% and close metrics (0.9781, 0.9766). DINO holds a high precision and recall and achieves 98.68% accuracy with DINO, closely followed by Moco v3 with a precision and recall yield of 99.48% accuracy but slightly behind SimCLR. In linear evaluation, SimCLR leads again with 91.04% accuracy, while BYOL is a close second with 85.94% (only a drop of 3.52%) and DINO sharply drops to 45.24%. MoCo v3 does comparably well at 94.84% In general, the most consistent model in both settings is SimCLR.

**Table 4** Fine-Tuning and Linear Evaluation Performance Metrics of SSL Models

| Fine-Tuning | | | | | Linear Evaluation | | | | |
|---|---|---|---|---|---|---|---|---|---|
| Model | Accuracy | Precision | Recall | F1 Score | Model | Accuracy | Precision | Recall | F1 Score |
| **SimCLR** | **99.64%** | **0.9964** | **0.9964** | **0.9964** | SimCLR | 91.04% | 0.9149 | 0.9104 | 0.9099 |
| BYOL | 97.66% | 0.9781 | 0.9766 | 0.9766 | BYOL | 85.94% | 0.8996 | 0.8594 | 0.8621 |
| DINO | 98.68% | 0.9870 | 0.9868 | 0.9868 | DINO | 45.24% | 0.5571 | 0.4522 | 0.4072 |
| Moco v3 | 99.48% | 0.9949 | 0.9948 | 0.9948 | **Moco v3** | **94.84%** | **0.9520** | **0.9484** | **0.9483** |

Evaluation metrics for four self-supervised learning models in SimCLR, BYOL, DINO, and Moco v3 are listed in Table 5. The results of testing show that SimCLR has the highest testing accuracy of 97.27% along with a precision, recall and F1 score of 0.9738,0.9726 and 0.9727, respectively. BYOL tails behind at 96.14%, and DINO and Moco v3 are further back at 95.58% and 94.92%, respectively. On validation, SimCLR achieves again the best score with 97.65% accuracy and the highest scores across all metrics Table 5 walks BYOL to 93.5%, DINO to 94.26%(less than Moco v3 95.86%). In general, and overall, SimCLR is the most reliable model for brain tumor classification.

Figure 6 displays class-by-class confusion matrices for the four fine-tuned SSL backbones. SimCLR has the least noisy matrix, DINO and Moco v3 both have distinct diagonals with lone misclassifications gathered on sparse borderline classes. BYOL has the most error dispersion with many off-diagonal assignments. To reduce these errors, consider multi-sequence fusion, sequence-aware augmentation, class-balanced sampling or focal loss, and post-training calibration or targeted re-labeling of ambiguous cases. SimCLR (or an ensemble of top backbones) is recommended for robust deployment.

**Table 5** Evaluation Metrics of Testing and Validation

| Testing | | | | | Validation | | | | |
|---|---|---|---|---|---|---|---|---|---|
| Model | Accuracy | Precision | Recall | F1 Score | Model | Accuracy | Precision | Recall | F1 Score |
| **SimCLR** | **97.27%** | **0.9738** | **0.9726** | **0.9727** | **SimCLR** | **97.65%** | **0.9769** | **0.9765** | **0.9762** |
| BYOL | 96.14% | 0.9646 | 0.9614 | 0.9611 | BYOL | 93.50% | 0.9395 | 0.9352 | 0.9340 |
| DINO | 95.58% | 0.9576 | 0.9558 | 0.9561 | DINO | 94.26% | 0.9465 | 0.9425 | 0.9430 |
| Moco v3 | 94.92% | 0.9535 | 0.9493 | 0.9494 | Moco v3 | 95.86% | 0.9601 | 0.9585 | 0.9588 |

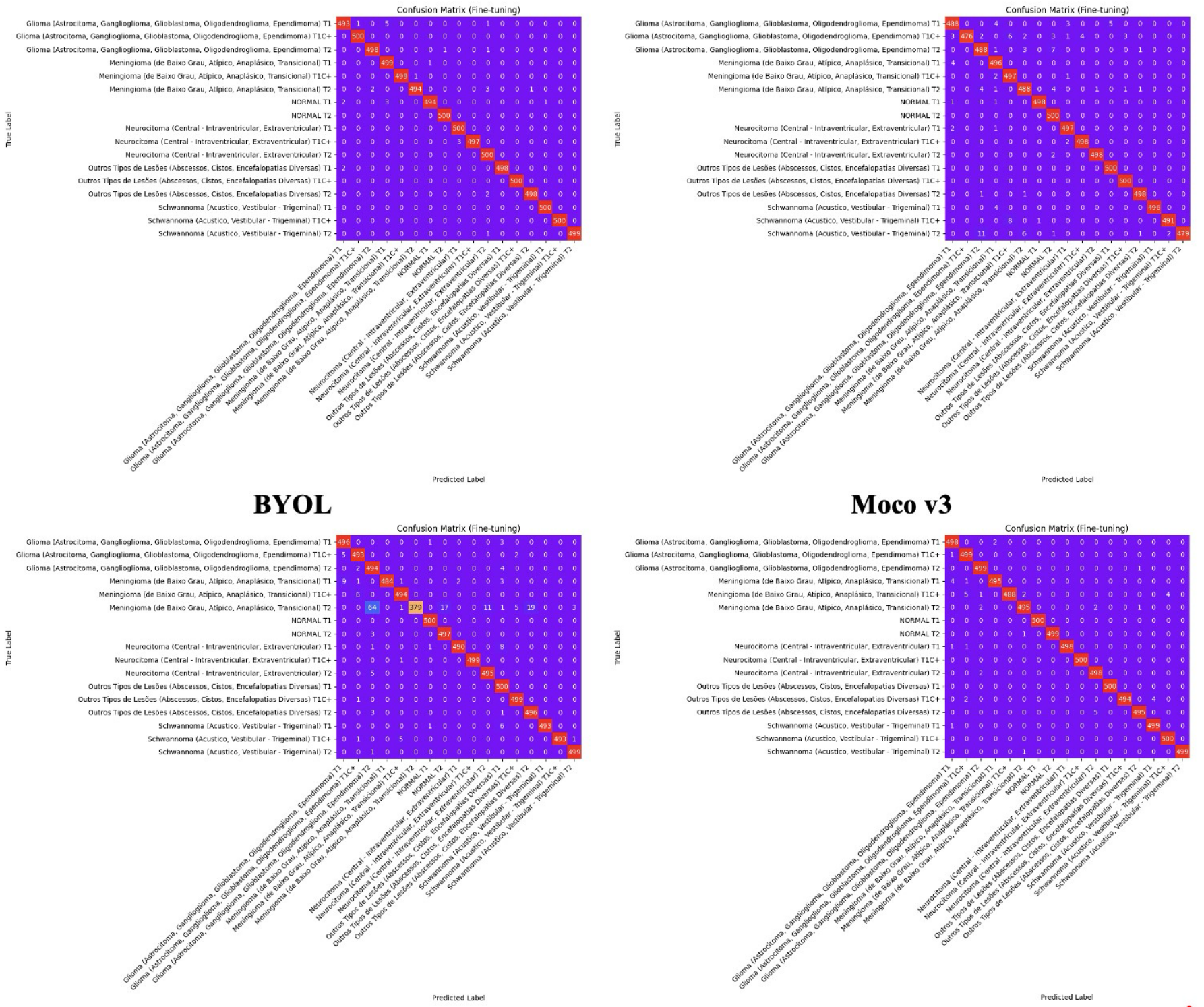

**Fig. 6** Confusion Matrix on Fine-tuning

Figure 7 plots fine-tuning training and validation losses for SimCLR, DINO, Moco v3 and BYOL. All four backbones show clear downward trends confirming effective learning and convergence yet their stability and evaluation noise differ. SimCLR produces the smoothest and steady train–validation gap, indicating the best generalization. DINO and Moco v3 both reduce loss rapidly early. BYOL attains low training loss but displays episodic, larger validation spikes.

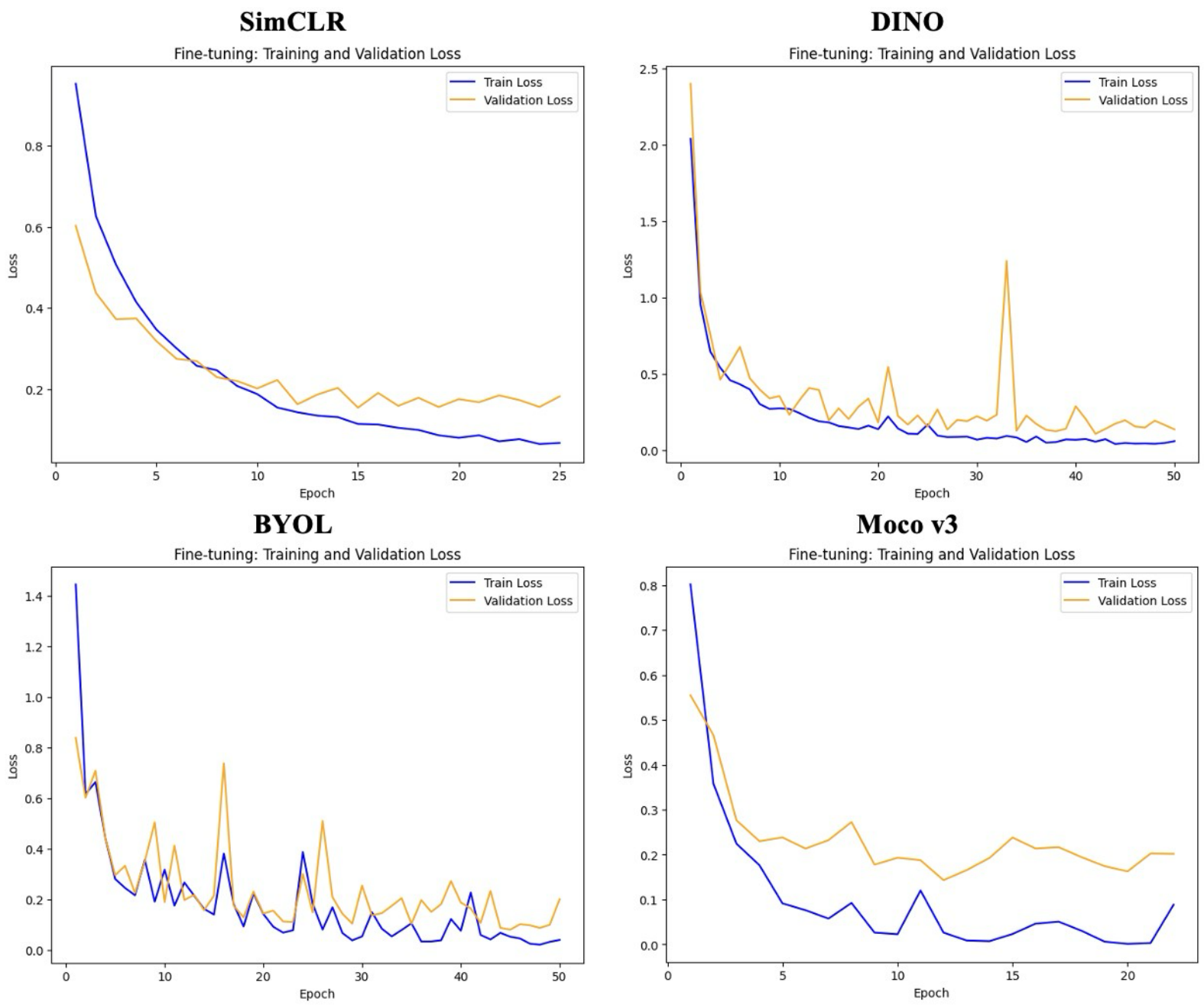

**Fig. 7** Training and Validation Loss on Fine-tuning

The t-SNE visualizations, as shown in Figure 8, further support that SSL features disentangle the learned representations, showing well-defined clusters across the learned representation for different tumor subtypes. In particular, the best visualization of t-SNE was obtained from the SimCLR approach, highlighting its ability to capture discriminative features among tumor classes.

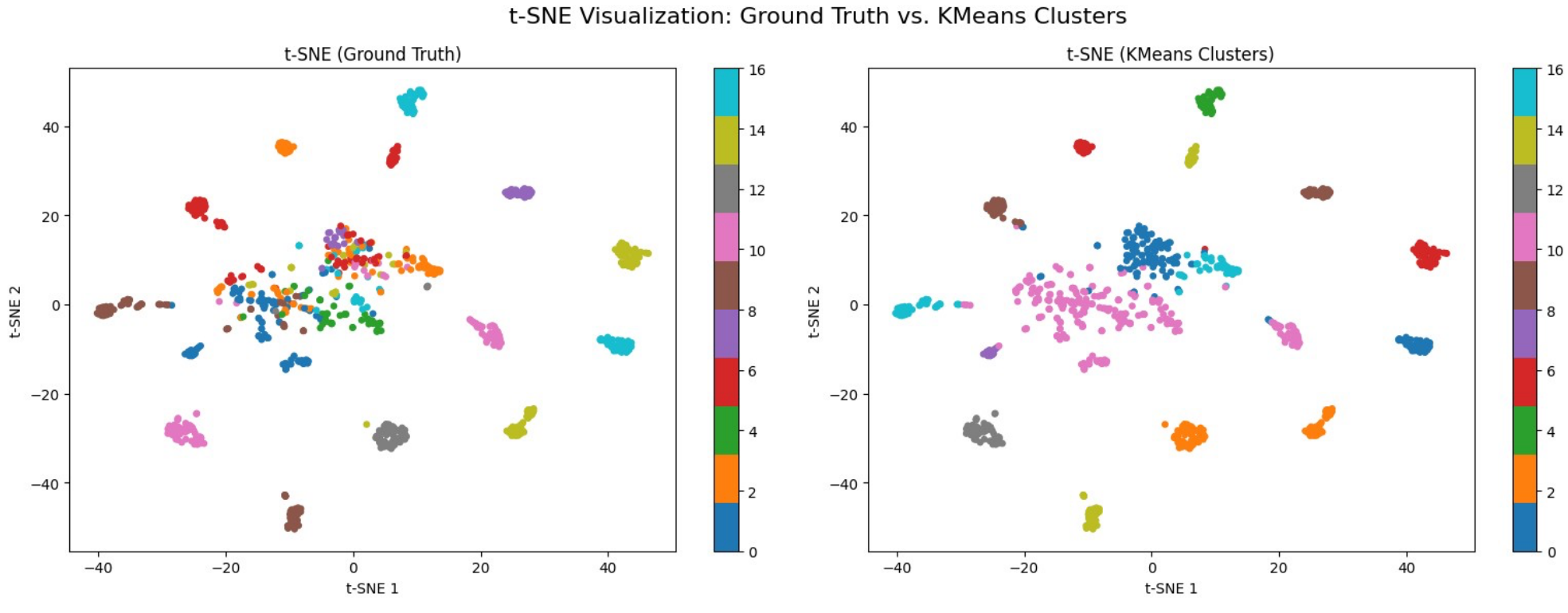

**Fig. 8** t-SNE Visualization: Ground Truth vs KMeans Clusters

## 4.2 Explainability of XAI Module

We employed explainable-AI methods, specifically Grad-CAM, Grad-CAM++, and Eigen-CAM as shown in Figure 9 to provide insight into the rationale underlying the model predictions. The techniques generate heat maps within the input images with emphasis on the locations that the network was basing its decisions on: Grad-CAM presents class-sensitive relevance by taking advantage of gradients, Grad-CAM++ generates higher resolution maps for multiple or small targets, and Eigen-CAM directly extracts saliency from feature activations.

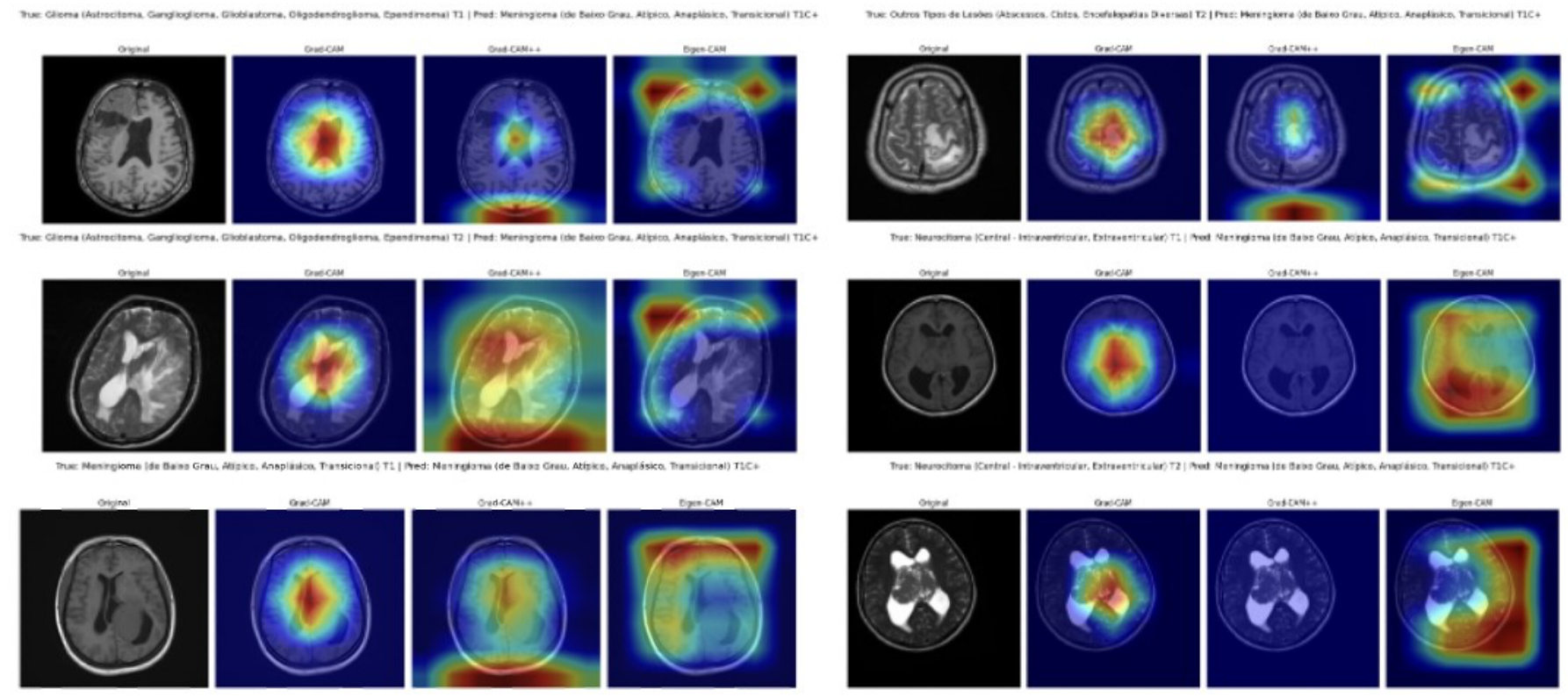

**Fig. 9** Grad-Cam, Grad-Cam++ and Eigen-Cam plot

# 5 Discussion

SimCLR outperformed four self-supervised learning (SSL) backbones—BYOL, DINO, and Moco v3—in brain MRI classification across various metrics, including accuracy, precision, recall, and F1 score. This success is attributed to its use of contrastive learning combined with effective data augmentation, allowing it to learn discriminative features essential for identifying subtle tumor subtypes. The model exhibited a more separable representation space, as shown by t-SNE clustering, which is crucial for distinguishing complex intra-class differences. Furthermore, a confusion matrix analysis indicated that SimCLR had minimal misclassifications compared to other methods. The explainability module (using Grad-CAM techniques) provided insights into clinically relevant tumor regions, enhancing model trustworthiness. Overall, SimCLR's robust learning approach and reliable discrimination make it the most appropriate SSL backbone for brain MRI classification in this study.

Table 6 highlights the advancements in brain MRI classification and tumor analysis, comparing previous methodologies to our approach. Earlier unsupervised frameworks like BYOL achieved 97.66% accuracy, while our method, utilizing virtual labeling and self-learning, reached 99.7%, showcasing excellent generalization. Classical methods such as decision trees achieved 96.2% accuracy. In segmentation, Swin UNETR outperformed U-Net with 96.27% accuracy, and the transformer-based DINOV2 excelled in image classification with nearly perfect results at 99.84% accuracy. Few-shot detection methods like CONSULT reached 93.67% accuracy. In tumor clustering, a combination of self-supervised learning and chemical imaging in the VQSRS framework achieved F1-score of 99.27%. Our SimCLR-based SSL approach demonstrated superior performance with 99.64% across accuracy, precision, recall, and F1-score, proving its robustness in multi-class brain MRI classification.

**Table 6** Comparison with Previous Studies on Brain MRI Classification and Tumor Analysis

| Task / Study | Best Model | Dataset | Accuracy (%) | Notes / Strengths |
|---|---|---|---|---|
| Self-Supervised MRI | BYOL (ResNet-50) | MRI dataset | 97.66 | Fine-tuned SSL backbone, strong representation learning |
| MRI Sequence Classification | MLvirtual | Hospital + Multicenter | 99.7 | Rule-based virtual labeling + selflearning, high generalization |
| Classical ML | Decision Tree | MRI features | 96.2 | High accuracy with 30-fold CV, simple interpretable model |
| Brain Tumor Segmentation | Swin UNETR | BraTS+CHGH | 96.27 | Outperforms U-Net, better for large tumors, uses self-supervised pretraining |
| Image Classification | DINOV2 ViT-S14 | Test set | 99.84 | Excellent handling of imbalanced data, perfect AUC |
| Few-Shot Tumor Detection | CONSULT | K2–K8 shots | 93.67 | Effective in low-shot settings, outperforms other anomaly detection methods |
| Chemical Imaging/ Tumor Clustering | VQSRS | SRH dataset | 88.22 | Self-supervised + proxy task; excels in clustering; captures latent tumor similarities; UMAP visualization; reconstructs chemical contrast |
| **This Study (2025)** | **SimCLRbased SSL** | **Multi MRI dataset** | **99.64** | **Best cross-domain generalization, robust for brain MRI classification** |

# 6 Conclusions

In this study, four self-supervised models for multi-class brain tumor imaging SimCLR, MoCo v3, DINO, and BYOL were analyzed. SimCLR emerged as the best performer across training, validation, and independent test sets, excelling in accuracy, precision, recall, and F1 score, while also showing a stable optimization process and clear tumor class separation. However, performance dropped during fine-tuning and independent tests, highlighting challenges like generalization, class imbalance, and overfitting. The single-sequence MRI focus may limit tumor characteristic representation. Future work will include multi-sequence and multimodal MRI data for improved representation, validation across diverse datasets, and exploration of advanced training methods to enhance generalization. This study reinforces SimCLR's effectiveness in automated brain MRI classification and outlines steps to increase trustworthiness in clinical applications.